# Dealing with Uncertainty in Situation Assessment: towards a Symbolic Approach

**Charles Castel, Corine Cossart, Catherine Tessier**
ONERA-CERT
BP 4025 - 31055 Toulouse Cedex 04 - France

## Abstract

The situation assessment problem is considered, in terms of object, condition, activity, and plan recognition, based on data coming from the real-word *via* various sensors. It is shown that uncertainty issues are linked both to the models and to the matching algorithm. Three different types of uncertainties are identified, and within each one, the numerical and the symbolic cases are distinguished. The emphasis is then put on purely symbolic uncertainties: it is shown that they can be dealt with within a purely symbolic framework resulting from a transposition of classical numerical estimation tools.

## 1 SITUATION ASSESSMENT PROBLEM

Let us consider the generic problem that is dealt with in the PERCEPTION project[1] [BCF+98], [CCMT97]: a symbolic representation of what is going on in an observed environment has to be built and updated, for applications such as surveillance, intelligence, or decision-aid systems, and autonomous systems. The environment includes mobile entities and is observed *via* various sensors (black and white, color and infrared cameras, radars). Numerical processings take sensor data as inputs and deliver recognized and tracked objects with symbolic properties (e.g. the type of the objects: *pedestrian, vehicle*...) and numerical attributes (position, speed...) The symbolic level interprets these objects in terms of on-going and future activities (e.g. *the pedestrian is going to take his car and leave the parking-lot*), so that the decision level (e.g. a human decision-maker) should be informed with semantically rich data and that further relevant actions should be undertaken.

Human observers may also be involved as "sensors", and their reports are direct inputs for the symbolic level.

---

[1] http://www.cert.fr/fr/dcsd/PUB/PERCEPTION/

## 2 PRINCIPLES

The set of the current activities is assessed by the symbolic level thanks to plan prototypes based on the expected properties and attributes of the objects and on the expected variations of the properties and attributes with time. Three basic notions are used:

• a *condition prototype* is an expected property *a priori* qualifying the objects that are likely to be observed. Condition prototypes are expressed by atomic formulas built from predefined predicates, e.g. *(type x pedestrian), (speed x 4km/h), (getting-closer x y)* with $x$ and $y$ being variables;

• an *activity prototype* is a set of expected conditions and constraints *a priori* qualifying the objects that are likely to be observed. Activity prototypes are represented by constrained cubes [TGLP88], i.e. conjunctions of atomic formulas associated with constraints, in which all the variables are assumed to be existentially quantified [CCMT97].
Ex: $\{(type\ x\ pedestrian),\ (type\ y\ vehicle),\ (speed\ y\ v)\ \{v = 0\},\ (getting\text{-}closer\text{-}to\ x\ y)\}$, with $x$, $y$ and $v$ being variables.

• a *plan prototype* is a temporal graph of activity prototypes; it is represented by an interpreted Petri net [DA91] whose places are associated with activity prototypes.
Ex:

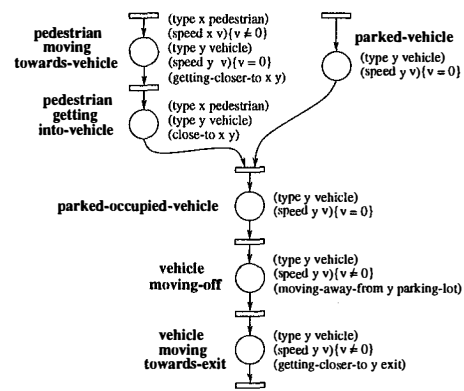

Figure 1: *vehicle-departure* plan prototype

Let $\mathcal{P}$ be the set of plan prototypes.

An *observation* $obs_n$ is a set of properties directly issued by the numerical processing, resulting from a numerical-symbolic translation, or issued by a human observer at time $t_n$. The *current situation* $S_n$ at time $t_n$ is a set of *plans* $(P_{i,m_i,n})$, defined as marked elements $P_i$ of $\mathcal{P}$; the marking $m_i$ of a plan prototype $P_i$ at time $t_n$ corresponds to the fact that some properties in $obs_n$ match the interpretation of some places (activity prototypes) in $P_i$.

Given $obs_{n+1}$ and $S_n$, the elaboration of the current situation $S_{n+1}$ at time $t_{n+1}$ is a prediction-verification process which is based on the following principles:
*i)* a greater importance is given to the continuation of existing plans; *ii)* all the objects appearing in properties within $obs_n$ have to be explained, i.e. to belong to at least one plan; *iii)* the prediction of situation $S_{n+1}$ from situation $S_n$ is the set of the reachable markings $m_i + k$ of plans $(P_{i,m_i,n})$; *iv)* the verification consists in matching the properties of $obs_{n+1}$ with those reachable markings; if some properties remain unmatched this way, new plans $(P_{j,m_j,n+1})$ are created.

As a given object may be associated with several plans, $S_{n+1}$ represents the different hypothetic plans that are likely to be in progress in the observed environment.

## 3 UNCERTAINTY ISSUES

Whatever the situations are built for (immediate or delayed warnings or actions, information collecting in an intelligence context, detection of specific activities...), the situation assessment process has to deliver appropriate results, which means that [KSH91]:
- results (assessed situations) have to agree with the global mission goal: a potentially hazardous situation has to be reported early, even if the assessment is not complete or certain; all the situations that are significant for the mission must be expected and recognized.
- results have to be accurate, i.e. situations must not include a high number of different plan hypotheses. Therefore, activity and plan prototypes, as well as the matching algorithm, have to be discriminating enough (a plan prototype that would model that anything can happen is of minor interest).
- results have to be computed efficiently, without too numerous or too complicated models.

Let us now analyze the situation assessment process from the uncertainty point of view.

### 3.1 UNCERTAINTY AND THE MODELS

The whole situation assessment process is a series of transformations from sensor data into high level symbolic properties.

#### 3.1.1 Conditions

Conditions are the first link between sensor data and the symbolic level. They include:

- numerical attributes, such as the *position* in the environment, or the *speed*, of a tracked object;
- classification results, such as the *type* of the objects (e.g. an object is classified as a *pedestrian* or a *vehicle*);
- elementary actions that may result from a numerical-symbolic translation (e.g. *getting-closer-to, close-to*).

The accuracy of the numerical attributes depends on the numerical processings and is only a matter of numerical precision. On the other hand, classes and elementary actions have to be *a priori* defined, and the accuracy of the definition has direct consequences on object matching: a compromise has to be found between too strict and too loose definitions. For example, condition *getting-closer-to* will hold for a pedestrian *P1* moving towards a parked vehicle *V1* if P1's speed vector belongs to a given cone (see figure).

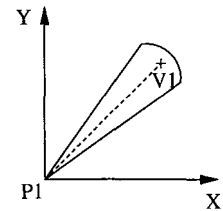

Figure 2: *getting-closer-to(P1, V1)*

This kind of definition allows the dispersion of natural behaviours and the imprecision on the numerical measurements to be taken into account. Nevertheless, there is a threshold effect (the condition holds if *P1* is within the cone and does not hold even if he is close to it). Furthermore, a unique definition may not be suited to a real environment in so far as actual behaviours for achieving a given goal may be very different from one another, depending on the particular objects at stake and on the environmental context (weather, environment layout...)

The second point is that a given condition may include several sub-conditions (this is the case for class conditions that are most of the time defined by several parameters): should all the sub-conditions hold to make the condition hold too? or are there any sub-conditions that are less important than the others? what is the difference for condition assessment when a sub-condition is mismatched, and when it is not matched at all through lack of information?

#### 3.1.2 Activities

The assessment of an activity, as a set of conditions, results from the assessment of the conditions describing it and therefore inherits the corresponding uncertainty issues. Nevertheless further particular issues can be identified.

Activities are closely linked to the types of the objects as they are the *actors*. The uncertainty on types (e.g. an object is close to the *pedestrian* class, but also to the *vehicle* class) may result in several exclusive activity hypotheses being created (e.g. *pedestrian-moving* OR *vehicle-moving*).

An activity may not be fully assessed or distinguished from another one by means of its mere condition set. For instance, an empty parked vehicle cannot be distinguished from a parked vehicle with a driver inside: what can be told is that there is a *parked vehicle*. A refinement can be made thanks to the current plan hypotheses and the history of the observations: if one of these hypotheses is *vehicle-departure* plan and a *pedestrian* just disappeared near the *vehicle* involved, activity *parked-occupied-vehicle* is most likely to hold.

Finally, the same questions as before are raised: should all the conditions hold to make the activity hold too? or are there any conditions that are less important than the others? what is the difference for activity assessment when a condition is mismatched, and when it is not matched at all through lack of information?

### 3.1.3 Plans

The assessment of a plan, as a temporal sequence of activities, results from the assessment of the activities and therefore inherits the corresponding uncertainty issues. A particular problem that arises when considering "free" environments, where no predefined procedure is available, is the assessment of the temporal sequence of the activities: does the set of the assessed activities match the temporal sequence described in the prototype? are other sequences acceptable? what to do with extra activities?

### 3.2 UNCERTAINTY AND MATCHING

As an explanation is required for each observed object, a given object has to be matched to one activity within one plan. What happens as a result of the prediction - verification process is that this object may be matched to *several* activities and plans, and conversely, a given plan may be associated to several different objects. The questions that are raised are the following: should all the matchings be kept or not? should the plan hypotheses with no new corresponding observations be kept and for how long, given that there may be objects that are occulted or not observable? are there hypotheses that are more relevant than the others, considering the mission?

## 4 CLASSIFICATION OF UNCERTAINTIES

The previous analysis results in a classification of the uncertainties within the situation assessment problem. Three main types of uncertainties may be identified: matching uncertainties, conjunction uncertainties, and disjunction uncertainties.

### 4.1 MATCHING UNCERTAINTIES

Matching uncertainties are linked to the data-to-model matching problem: data coming from sensors or issued by numerical processings have to be matched with predefined models corresponding to higher semantic levels, e.g. the *pedestrian* type, the *getting-closer-to* condition, the *vehicle-moving* activity, or the *vehicle-departure* plan. As stated before, data hardly perfectly match predefined models, and *imperfect matchings* have to be considered. Nevertheless, two different cases have to be distinguished:

• a first case is *numerical* imperfect matching, which involves parameters that take their values in *continuous* domains, where distances can be defined.

Ex: *(i)* the position or speed associated to an object is imprecise, depending on the sensor data or the numerical algorithms.

A traditional approach for the estimation of such parameters is Kalman filtering [MM93], which allows the state (i.e. the set of the parameters such as position and speed) of a dynamic system to be predicted and updated by new measurements, on the basis of a quadratic error criterion. Noise effects coming from both measurements and modelling are taken into account. It is widely used in object dynamic tracking (e.g. [JKC97]).

*(ii)* condition *getting-closer-to* is more or less satisfied, depending on the position of the pedestrian with respect to the cone (see figure 2).

This is classically dealt with thanks to probabilistic or possibilistic approaches, depending on the available knowledge. For example, Herzog, in the VITRA project [Her95], uses a measure of degrees of applicability that expresses the extent to which a spatial relation is applicable. Every relevant geometric factor (relative distance, angular deviation from a given canonical direction...) is mapped onto interval [0, 1] by means of cubic spline functions associated with different qualitative notions (such as the contiguity or the proximity for the relative distance). The degree may depend on different spline values, e.g. the degree of relation *right of* is a combination of the direction and proximity factors.

In a traffic context, [FHKN96] use predicates to describe the (relative) motion of one (or two) objects. Primitives such as *fast(X, t), equal-orientation(X, Y, t)*, which can be directly derived from the speed and orientation attributes estimated by the tracking process, are modelled by means of fuzzy sets.

• a second case is *symbolic* imperfect matching, which involves symbolic items, i.e. discrete frames, within which no distance can be defined.

Ex: *(i)* property *(type V1, car)* imperfectly matches prototype condition *(type y, bus)*; *(ii)* so does property *(type V1, car)* with prototype condition *(type x, pedestrian)*.

Hints to deal with such imperfect matchings will be given in the sequel.

It is worth noticing however that many symbolic items are simply abstractions of numerical features, especially when data only come from physical sensors: an object *V1* is labelled as a *car* because the values of the numerical parameters of the corresponding shape in the images (e.g. surface, position of the center of gravity, elongation, rotundity...) match the reference values of class *car*. Therefore, a numerical distance between e.g. a *car* and a *bus* or a *pedestrian* can be soundly defined as an aggregation of the res-pective distances between each parameter, thus allowing a matching quality coefficient to be defined. This projection onto a numerical space may also be propagated to the upper symbolic levels provided all the items involved have numerical bases. For example, in the static scene interpretation system described in [LLMC96], each object type is characterized by different attributes (geometric attribute, aspect attribute...) The validation of the object hypotheses is based on the assessment of a global confidence degree for each object type. This global degree is a combination of the confidence degrees of the different attributes, which are directly computed from numerical characteristics of the detected shapes in the images.

Purely symbolic items do not have any numerical bases and therefore cannot be projected onto any numerical space *without adding any supplementary knowledge*, as weights, preferences, etc. Examples of such symbolic items are data coming from human observers, and condition, activity and plan prototypes.

## 4.2   CONJUNCTION UNCERTAINTIES

Conjunction uncertainties arise when several sub-conditions, conditions or activities have to hold to make a condition, activity or plan respectively hold.

• In the case where numerical matching quality coefficients can be defined for each component of the item to be assessed, a common approach is to aggregate them, following given rules that most often depend on external knowledge. In [DP95], Dubois and Prade extend the basic principles of fuzzy pattern matching (characterized by data which can be pervaded by imprecision and uncertainty, and requirements which may be fuzzy) to situations where different subparts of a pattern have various levels of importance. They develop the case where the importance weight becomes a function of the concerned attribute value. In [BL96] it is noticed that, with classical likelihood aggregation rules such as min/max in the possibility theory, the matching quality decreases as the description is more detailed, because of the imperfect matching of individual details. Therefore the concept of description redundancy is defined, which allows the matching likelihood to be assessed by selecting a limited number of description items. A method is described to assess how many items, and which ones, may be dropped.

• In the other cases, there is no means to *quantify* to what extent the set of components matches the item.
Ex: *(i)* let *{(type y vehicle), (speed y 15km/h)}* be a two-condition activity prototype. This activity holds if there exist an object *y* assessed as a vehicle with a 15km/h speed. Let us suppose that the lower level (or a human observer) issues the following data: *{(speed O1 10km/h), (close-to O1 building)}*, which means that an object *O1* with a speed of 10km/h was detected near the building. To what extent does this observation match the activity prototype? "0.5", as only one predicate can be matched out of two? or more, as *10km/h* is quite close to *15km/h*? how is it possible to quantify the missing condition *(type y vehicle)* and the additional one *(close-to O1 building)*?
*(ii)* to what extent does activity sequence (*pedestrian-moving-towards-vehicle, pedestrian-stopped, pedestrian-moving-towards-vehicle, pedestrian-getting-into-vehicle, vehicle-moving-towards-exit*) match plan prototype *vehicle-departure*?
Hints to deal with such symbolic conjunction uncertainties will be given in the sequel.

## 4.3   DISJUNCTION UNCERTAINTIES

Within the prediction - verification process, a set of disjunctive hypotheses of conditions, activities or plans is created each time several matchings are possible between data and models: therefore, hypothesis disjunction uncertainties are a consequence of matching and conjunction uncertainties described above.

• In the case where numerical matching quality coefficients can be associated to conditions and activities, and likelihood coefficients can be associated to the transitions between activity prototypes within plan prototypes, each disjunction element can be qualified by a global coefficient resulting from the aggregation of the different matching quality and likelihood coefficients. If pruning is necessary, a preference function based on these global coefficients or on external knowledge can be defined. For example, [HB93] propose a task-driven approach to the surveillance problem in traffic, characterized by a selective attention. A dynamic form of Bayesian network is used to capture the changing relationships between scene objects: given a task (e.g. *attend to likely overtaking and ignore likely following*), it provides measures of which pairs of objects are worth further attention. Then a dedicated tasknet is used to identify the likelihood of the wanted task (e.g. *overtaking*), of the related but unwanted tasks(e.g. *following, queueing*), and of the default unknown task. Tasknets are static Bayesian networks, with *a priori* conditional probabilities that reflect a preferential bias towards a feature that is deemed to be most interesting.

• In other cases, there is no means to *quantify* to what extent a plan hypothesis is better than another one, except with external knowledge based criteria, such as mission dependent preferences. In [CG91], a system for story understanding is described in which Bayesian networks are dynamically constructed in order to evaluate the conditional probabilities of competing plan hypotheses given the evidence. The prior probability of each hypothesis is computed under the assumption of a large but finite domain of equiprobable elements: the probability is linked with the number of instances of the plan. [Bau94] presents a framework based on Dempster-Shafer's theory for assessing and selecting plan hypotheses that takes into account disjunctive and uncertain observations as well as agents' preferences. Agent-specific preferences are encoded as basic probability assignments. A total ordering of the hypotheses can be ob-tained by collapsing the belief intervals computed for each hypothesis into a pignistic probability.

## 5 SYMBOLIC UNCERTAINTIES

In this section, we particularly focus on purely symbolic items for which no intrinsic numerical basis is available. Symbolic data may be projected onto numerical spaces as, in some cases, predefined likelihood or preference measures encode notions such as sensor reliability, information quantity or matching satisfaction. But, as it is a matter of context, no universal method is available [DP94]. What is aimed at is to show that, within the situation assessment framework, some notions widely used in continuous contexts can be transposed within discrete ones.

### 5.1 AN ESTIMATION PROBLEM

The prediction - verification principle explained in section 2 shows that the predicted state of the observed environment (i.e. the activities to come) is computed from previous information and models (plan prototypes), and revised with new information, within a dynamic process: it is the same principle as numerical estimation. Therefore, dynamic situation assessment, at the activity and plan levels, is a *symbolic estimation problem*.

Existing estimation techniques are numerical techniques: they aim at assessing deterministic or random magnitudes from observations tainted with stochastic errors. Kalman filtering, already mentioned in section 4.1, is one of these techniques. As our problem is to estimate the situation in a dynamic environment from uncertain reference models on the one hand and uncertain data on the other, the idea is to reconsider the symbolic layers of situation assessment in the light of numerical estimation techniques. The basic notions of Kalman filtering are especially investigated, and adapted to the symbolic context of situation assessment.

### 5.2 FROM KALMAN FILTERING TO SYMBOLIC ESTIMATION

Kalman filtering addresses dynamic systems whose state equation involves a matrix representing how the state varies with time, a deterministic input, and a random noise, the state noise, modifying the deterministic evolution of the state. The observation equation links the current observation to the current state via a second matrix and a second type of random noise, the observation noise.
*Prediction* of the state estimate at time $t_{n+1}$ is accomplished from the state estimate at time $t_n$, so is the predicted observation at time $t_{n+1}$. The covariance matrix of the state estimation error is also computed.
The second step consists in comparing the observation at time $t_{n+1}$ with the prediction, and consequently to *revise* the state estimate. This revision depends on the prediction errors and on the noises.

The notions and principles of Kalman filtering are now going to be transposed into the symbolic world.

#### 5.2.1 Symbolic state

In Kalman filtering, the state is assumed to be a gaussian variable characterized by a mean and a covariance. This can be transposed through the notions of kernel plan and kernel activity, and plan and activity tolerance, thus redefining the plan and activity prototypes.

**Definition 1** *a plan prototype $P_i$ is a pair $(KP_i, T(P_i))$. The kernel plan $KP_i$ is a minimum sequence of activity prototypes that has to be* necessarily *matched by a sequence of observations in order to interpret them as an instance of $P_i$. The plan tolerance $T(P_i)$ is a dispersion around $KP_i$: it is a set of supplementary activity prototypes that will* possibly *be matched by the observed sequence.*
Let $\mathcal{P}$ be the set of plan prototypes.

Ex: kernel plan of *vehicle-departure* plan prototype is sequence:

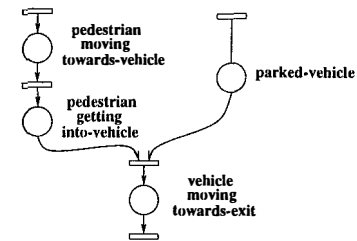

Figure 3: kernel plan for *vehicle-departure*

Plan tolerance may include *pedestrian-stopped* and *vehicle-moving-off* activity prototypes.

**Definition 2** *an activity prototype $A_{ij}$ within a plan prototype $P_i$ is a pair $(KA_{ij}, T(A_{ij}))$. The kernel activity $KA_{ij}$ is a minimum set of conditions and constraints that have to be* necessarily *matched by an observation in order to assess it as an instance of $A_{ij}$. The activity tolerance $T(A_{ij})$ is a dispersion around $KA_{ij}$: it includes both numerical dispersions around conditions and constraints of $KA_{ij}$, and supplementary conditions and constraints, that will* possibly *be matched by the observed activity.*

Ex: kernel activity of *moving-vehicle* activity prototype is *{(type y vehicle) (speed y v)}*; activity tolerance may include *{v ≥ 30km/h}* constraint or *(speed y 25km/h)* condition (numerical dispersion), and supplementary symbolic conditions such as *(make y Renault), (moving-backwards y)*.

**Definition 3** *the current state $S_n$ at time $t_n$ is a set of marked plan prototypes of $\mathcal{P}$, i.e. a set of plans $(KP_{i,m_i,n}, T(P_{i,m_i,n}))$. For each plan, $KP_{i,m_i,n}$ is the Petri net associated to kernel plan $KP_i$ and marked with marking $m_i$ at time $t_n$; $T(P_{i,m_i,n})$ is the set of supplementary activities belonging to $T(P_i)$ that hold at time $t_n$.*

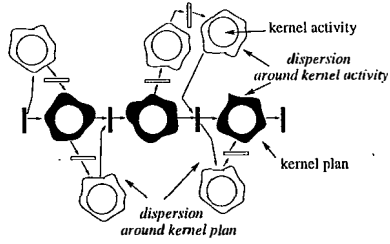

Figure 4: kernel activities, kernel plan, and dispersions

*Marking $m_i$ corresponds to a subset of activity prototypes, i.e. pairs $(KA_{ij,m_i,n}, T(A_{ij,m_i,n}))$: it means that each $KA_{ij,m_i,n}$ holds at time $t_n$ and that some conditions of $T(A_{ij})$ also hold. It is the same for supplementary activities within $T(P_{i,m_i,n})$.*
*The equation of evolution for state $S_n$ is given by the Petri net based plan prototypes, since they represent the temporal linkings of the activities. The inputs are events associated to the transitions of the Petri nets (e.g. event* (type x pedestrian) *meaning that a pedestrian is appearing, thus modifying the symbolic state of the system).*

The main difference with the numerical case is that the prediction temporal horizon is finite for one state, since plan prototypes involve finite sequences of activities.

#### 5.2.2 Symbolic observations

Let us assume that only purely symbolic observations are available, i.e. data coming from human observers or not assessed with numerical criteria.

**Definition 4** *as for the numerical case, the observation equation is a projection of the current state onto the set of actually observable activities. This projection depends on observation conditions (environment layout, observation means).*

#### 5.2.3 Noises

As for the numerical case, two different kinds of noises can be distinguished: state noise and observation noise, respectively modifying the state and observation equations.

**Definition 5** *a state noise is a deviation of a plan from the plan prototype, resulting from unexpected events created by objects that do not belong to that plan prototype.*

Ex: *(i)* a dog may cross the road just ahead of a car leaving the parking-lot; this car has to brake suddenly, thus modifying plan *vehicle-departure*; *(ii)* two independent plans may interact: the *vehicle-departure* car may have to brake because of a *vehicle-arrival* parking car.

**Definition 6** *an observation noise is a deviation of the actual observation from what should be observed given the current state.*

Ex: *(i)* dysfunctions within the observation means, or a bad weather, may alter the observations; *(ii)* objects that have nothing to do with the on-going plans are observation noises for these plans, e.g.: a dog wandering on the parking-lot; objects belonging to several independent and non-interacting plans are observation noises for one another; moreover, these objects may create unexpected occultations.

Remark: observation noise does not modify the current state as it is just a matter of perception conditions, whereas state noise does. It is worth noticing however that an observation noise may become a state noise, e.g. the wandering dog may cause the *vehicle-departure* car to brake suddenly. In that sense, both kinds of noises can be correlated, as it may be the case in Kalman filtering.

### 5.3 DYNAMIC ASPECTS

#### 5.3.1 One-step prediction

**Definition 7** *let $S_{n+1|n}$ be the one-step predicted situation from $S_n$. $S_{n+1|n}$ is the set of the reachable markings $m_i + k$, $k \in \{0, 1\}$ of plans $(KP_{i,m_i,n}, T(P_{i,m_i,n}))$: it is a disjunction of activity prototype subsets $\{(KA_{ij, m_i+k, n+1|n}, T(A_{ij, m_i+k, n+1|n}))\}, k \in \{0, 1\}$.*

Ex: let us assume that situation $S_n$ corresponds to activity subset *{pedestrian-moving-towards-vehicle, parked-vehicle}* within plan *vehicle-departure*. $S_{n+1|n}$ is the disjunction *{pedestrian-moving-towards-vehicle, parked vehicle}* OR *{pedestrian-getting-into-vehicle, parked-vehicle}*.

Remark: $T(A_{ij, m_i+1, n+1|n})$ is the expected dispersion around the predicted activities associated with marking $m_i + 1$; it is composed of propagated matched conditions of $T(A_{ij, m_i, n})$.
Ex: if *(make y Renault)* is a supplementary condition of activities *pedestrian-moving-towards-vehicle* and *parked-vehicle*, it is propagated by the prediction process and associated to the disjunction *{pedestrian-moving-towards-vehicle, parked vehicle}* OR *{pedestrian-getting-into-vehicle, parked-vehicle}*

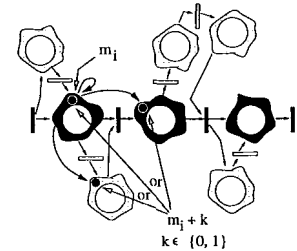

Figure 5: prediction process

**Definition 8** *the predicted observation, denoted $obs_{n+1|n}$, results from the projection of $S_{n+1|n}$ onto the observable space. $obs_{n+1|n}$ is therefore the disjunction of observable activity prototype subsets*

$\{(K^{obs}A_{ij,\ m_i+k,\ n+1|n}, T^{obs}(A_{ij,\ m_i+k,\ n+1|n}))\}$,
$k \in \{0, 1\}$.

### 5.3.2 Revision

A main notion in Kalman filtering is innovation, which represents the observation prediction error, i.e. the difference between predicted and actual observations. Innovation is the basis for state estimate revision. In the same way, the notion of symbolic innovation can be defined.

**Definition 9** *the symbolic innovation $I_{n+1}$ results from the matching of observation $obs_{n+1}$ delivered at time $t_{n+1}$ with predicted observation $obs_{n+1|n}$. Revised state $S_{n+1|n+1}$, denoted $S_{n+1}$, is a function of $I_{n+1}$.*

Two cases have to be distinguished:
- $obs_{n+1}$ matches $obs_{n+1|n}$ (i.e. one subset of observable activity prototypes within the predicted disjunction); what is observed instantiates what is expected, therefore, state estimate $S_{n+1}$ is within prediction $S_{n+1|n}$, there is no innovation.

Ex: if $obs_{n+1|n}$ is *{(type P1 pedestrian), (speed P1 v)}* (predicted kernel activity) with tolerance *{v ≤ 8km/h}* within a *pedestrian-moving* plan, and $obs_{n+1}$ is *{(type P1 pedestrian), (speed P1 5 km/h)}*: $obs_{n+1}$ is more precise than $obs_{n+1|n}$ because of the speed value. This numerical dispersion is included within the predicted activity tolerance.

- $obs_{n+1}$ does not match $obs_{n+1|n}$. An *imperfect matching*, characterizing the common features of $obs_{n+1}$ and $obs_{n+1|n}$ is then considered. Let $inf_{n+1}$ be the result of this imperfect matching.

1. if $inf_{n+1}$ matches the kernels of one subset of the predicted activities and possibly some of the predicted tolerance conditions, innovation $I_{n+1}$ is the set of supplementary properties contained in $obs_{n+1}$. Revision then depends on these properties:
- if they involve objects already known or expected within the current plans, they are integrated as extended tolerances within the current plans.
Ex: if $obs_{n+1}$ is now *{(type P1 pedestrian), (speed P1 5 km/h), (close-to P1 building)}*, supplementary property *(close-to P1 building)* concerns object *P1* and can be integrated within plan *pedestrian-moving* tolerance; or, if a more specific plan prototype exists and includes this property, revision is a switch from the less specific to the more specific plan.
- if they involve objects that are unknown and unexpected, they are noises for the current plans. Revision consists in instantiating new plan prototypes in $\mathcal{P}$ involving these properties.
Ex: if $obs_{n+1}$ is now set *{(type P1 pedestrian), (speed P1 5km/h), (type V1 vehicle), (speed V1 25km/h)}* supplementary properties – which are observation noise for the current plan – are associated with a new object *V1*. Therefore, activity prototype *vehicle-moving* within *vehicle-arrival* plan prototype can be instantiated.
In this case, observation noises may also lead to instantiate new plan prototypes with fictitious objects. Only an improvement in observation conditions can discard those groundless hypotheses.
- a particular case occurs when supplementary properties link already existing and new objects: revision then consists in matching a more specific plan within $\mathcal{P}$.
Ex: let us assume that $obs_{n+1}$ is the previous set with supplementary property *(getting-closer V1 P1)*. The plan is not *pedestrian-moving* plan anymore but *car-picking-up-pedestrian* plan.

2. if $inf_{n+1}$ does not match the kernels of any subset of the predicted activities, revision consists in considering other plans within $\mathcal{P}$. Nevertheless, if some properties are common to $inf_{n+1}$ and to the kernels, they can direct the choice to more relevant plans.
Ex: let us assume that $obs_{n+1}$ is now *{(speed P1 v), (close-to P1 building)}*. $inf_{n+1}$ would be *(speed P1 v)*. Therefore, the on-going plan is necessarily an *object-moving* plan.

Remarks:
- The computation of imperfect matching $inf_{n+1}$ has already been studied for conditions expressed as logical cubes (with no constraints) [CCMT97]. It is based on cube anti-unification and reduction. The extension to constrained cubes is currently under study.
- For the time being, there is no *a priori* links between plan prototypes yet. A graph-based structure of those prototypes, with less specific - more specific links is currently being studied, so as a hierarchy of the involved objects.

## 6 DISCUSSION

Considering the purely symbolic part of situation assessment as an estimation problem brings several improvements to the principles set out in section 2.

The first point is that purely symbolic uncertainty is dealt with within a symbolic framework, without projecting it onto subjective numerical spaces. It has been shown that notions such as symbolic kernels and tolerances and symbolic noises could be defined. Hints towards symbolic imperfect matching are also given.

One consequence of the estimation approach is that the result issued is less combinatorial, since some of the uncertainties fall within noises or tolerances and no longer create new activities or plans. Moreover, it allows a least commitment strategy to be followed in so far as least detailed activities and plans are first selected for matching. In the same way, minimum changes are given greater importance at the plan level, just as for numerical estimation algorithms in which the inertia of past events is a basis for prediction. It is worth noticing that this point of view is different from [DDdSCP95]'s for example.

Nevertheless, four important remarks have to be made:
- Disjunction is a powerful tool when several items can no longer be considered within the same model: that is why it is used in Kalman filtering especially in multi-target tracking (multi-model Kalman filtering [BBS88]). Consequently, the symbolic noises and tolerances that are introduced do not eliminate disjunctions, but contribute to reduce their drawback, i.e. the combinatorial explosion.
- The previous remark is particularly important when "noise becomes signal". For example, let us suppose that plan *vehicle-departure* is going on in the environment, with current activity being *pedestrian-moving-towards-vehicle*. Another vehicle entering the parking-lot is state noise for this plan. But if this latter vehicle gets close to the pedestrian and the driver attacks the pedestrian, this is no longer noise, but a *mugging* plan : models have to be switched.
- The framework that is proposed is a first step towards activity and plan learning in so far as tolerances and noises allow unpredicted items to be intergrated within the models.
- Obviously, situation assessment problems are seldom purely symbolic (or purely numerical). As a matter of fact, a further approach would be to use both quantitative and symbolic handling of uncertainty in the same application [KSH91], i.e. to mix two *a priori* very different worlds...